\documentclass[conference]{IEEEtran}
\usepackage{enumerate}
\usepackage{booktabs}
\usepackage{multirow}
\usepackage{algorithm}
\usepackage{algpseudocode}
\usepackage{comment}
\usepackage{amsmath,amssymb}
\usepackage{latexsym}
\usepackage{graphicx}
\usepackage{subfigure}
\usepackage{epstopdf}
\usepackage{epsfig}
\usepackage{xspace} 
\usepackage[bottom]{footmisc}
\usepackage[11pt]{moresize}

\newcommand{\model}{\textbf{\texttt{EnHDC}}}
\newcommand{\UCIHAR}{\textbf{\texttt{HAR}}}
\newcommand{\CARDIO}{\textbf{\texttt{CARDIO}}}
\newcommand{\ISOLET}{\textbf{\texttt{ISOLET}}}
\newcommand{\MNIST}{\textbf{\texttt{MNIST}}}
\usepackage{url}
\newcommand{\mypara}[1]{
	\vspace*{0.01cm}
	\noindent\textbf{\textit{#1}}}

\begin{document}

\title{EnHDC: Ensemble Learning for Brain-Inspired Hyperdimensional Computing}
\author{
	Ruixuan Wang, Dongning Ma, Xun Jiao \\
	\{rwang8, dma2, xun.jiao\}@villanova.edu \\
	Villanova University \\
	Villanova, PA 19085
}

\maketitle
\begin{sloppypar}
\begin{abstract}
Ensemble learning is a classical learning method utilizing a group of weak learners to form a strong learner, which aims to increase the accuracy of the model.
Recently, brain-inspired hyperdimensional computing (HDC) becomes an emerging computational paradigm that has achieved success in various domains such as human activity recognition, voice recognition, and bio-medical signal classification. HDC mimics the brain cognition and leverages high-dimensional vectors (e.g., 10000 dimensions) with fully distributed holographic representation and (pseudo-)randomness. 
This paper presents the first effort in exploring ensemble learning in the context of HDC and proposes the first ensemble HDC model referred to as \model. 
\model ~uses a majority voting-based mechanism to synergistically integrate the prediction outcomes of multiple base HDC classifiers. 
To enhance the diversity of base classifiers, we vary the encoding mechanisms, dimensions, and data width settings among base classifiers.
By applying \model ~on a wide range of applications, results show that the \model ~can achieve on average 3.2\% accuracy improvement over a single HDC classifier. Further, we show that \model ~with reduced dimensionality, e.g., 1000 dimensions, can achieve similar or even surpass the accuracy of baseline HDC with higher dimensionality, e.g., 10000 dimensions. This leads to a 20\% reduction of storage requirement of HDC model, which is key to enabling HDC on low-power computing platforms.  

\end{abstract}

\section{Introduction}
Inspired by how human brain functions, hyperdimensional computing (HDC) is an emerging computing scheme that leverages the abstract patterns and mathematical properties of vectors in high dimension spaces~\cite{rahimi2016hyperdimensional, najafabadi2016hyperdimensional}. Rather than processing actual numbers, HDC works with hypervectors (HV), which are high dimensional (e.g., 10000 dimensions), holographic (not micro-coded) vectors with i.i.d. (independent and identically distributed) elements~\cite{kanerva2009hyperdimensional}. As a novel computing scheme, HDC has shown promising performance for various applications such as language classification~\cite{najafabadi2016hyperdimensional}, voice recognition~\cite{imani2017voicehd}, biomedical signal analysis~\cite{burrello2018one} and robotics~\cite{neubert2019introduction}.

Compared with traditional computing schemes such as neural networks, HDC has several advantages such as smaller model size and low computing cost, making it a promising computing scheme with low power computing platforms and edge computing devices~\cite{imani2019searchd}. In particular, training of HDC models is free of back-propagation and can be completed on a sample-basis instead of batches, making it superior to machine learning algorithms such as neural networks that requires intensive iterations of back-propagation to establish a model. This exposes more opportunities for one-shot learning and edge deployment. In addition, the memory-centricity of HDC grants the advantage of easily embracing the emerging energy-efficient in-memory computing schemes over other machine learning algorithms such as neural networks that are highly computation-centric~\cite{karunaratne2020memory}.  

Ensemble learning is a machine learning paradigm where multiple models (often called “weak learners”) are trained to solve the same problem and combined to get better results.
Typically, an ensemble learning system aims to improves the performance by combining diverse weak learners (base classifiers). In a practical problem setting, ensemble is useful for the following question: given a particular classification algorithm, which realization of this algorithm should be chosen? For example, for a multi-layer perceptron (MLP), different initializations and weights will render different decision boundaries. It is worth to note that choosing the best classifier with the smallest error in training data may not be necessarily the best classifier overall due to overfitting. Therefore, it remains a question as to which classifier should be chosen and one can be tempted to choose randomly with the risk of having a particularly poor model at last. Thus, using an ensemble model that combines the output from several models, e.g., averaging them, can reduce the risk of an unfortunate selection of a particularly poor classifier. 

For the first time, this paper explores the use of ensemble learning on HDC models and develop the first ensemble HDC classifier for a wide range of applications. The benefits of using ensemble learning on HDC includes improved accuracy and reduced model size.  
In particular, we make the following contributions: 
\begin{itemize}
    \item To the best of our knowledge, we propose the first ensemble HDC classifier called \model. By leveraging the aggregated intelligence of a variety of HDC classifiers with different random initializations, \model ~is able to achieve on average 3.2\% accuracy improvement over a single HDC classifier. 
    \item To further enhance the performance of \model, we improve the diversity of base classifiers by varying a diverse set of parameters for base HDC classifiers such as number of dimensions, data width, and encoding methods. This further leads to 1.2\% accuracy improvement over basic \model ~classifier.
    \item We evaluate the \model ~on four different practical application domains including image classification, human activity recognition, speech recognition, and medical diagnosis. \model ~enables HDC learning with smaller number of dimensionality, which leads to a 20\% model size reduction with no accuracy drop. 
\end{itemize}

\section{Related Work}
\mypara{Hyperdimensional Computing}
Most related works on HDC focus on two major directions: the application of HDC and the improvement of HDC processing. Particularly, HDC has been intensively applied into emerging applications such as robotics by integrating the sensory perceptions experienced by an agent with its motoric capabilities, which is vital to autonomous learning agents~\cite{mitrokhin2019learning}. In bio-sensing applications such as hand gesture recognition, HDC also shows 97\% accuracy, superior to traditional machine learning algorithms~\cite{moin2021wearable}.
HDC is also used in building efficient recommendation systems which is almost 14X faster and 7X energy efficient than traditional rating prediction systems~\cite{guo2021hyperrec}. For optimization of HDC, there are cross-layer in-memory or in-storage computing platforms designed for HDC to enhance energy efficiency~\cite{gupta2020thrifty}. Computation reuse opportunities are also explored to accelerate performance of HDC in FPGAs~\cite{salamat2020accelerating}. Software and hardware multi-fold approximation techniques in encoding are also applied to enhance efficiency ~\cite{khaleghi2020shear}.

\mypara{Ensemble Learning} Ensemble learning has been applied to various applications or scenarios for enhanced learning performance. In language translation tasks, transductive ensemble learning (TEL) is proposed to surpass marginal improvement on accuracy of traditional ensemble algorithms~\cite{wang2020transductive}. In zero-shot learning scenarios, multi-patch generative adversarial nets (MPGAN) with novel weighted voting strategies are also proposed for improvement of current ensemble learning algorithms for better performance~\cite{chen2020rethinking}. Ensemble learning is also introduced into transfer learning to mitigate the over-fitting drawbacks of current transfer strategies~\cite{zhong2020translider}. By parallelism and sharing weights information amongst members in the ensemble classifier, \textit{BatchEnsembles} achieves 3X speed-up and 3X memory reduction over traditional ensemble algorithms~\cite{wen2020batchensemble}.

\mypara{Our Work} Some earlier literature indicates having a group of classifiers each targeting at one feature to build an ensemble can achieve higher classification accuracy while maintaining lower memory footprint~\cite{burrello2018one}. However, a systematic analysis of ensemble learning performance with HDC is still absent. \model, to the best of our knowledge, is the first work that systematically explores ensemble learning performance under different configurations such as \textbf{encoding methods}, \textbf{dimensions of HV} and \textbf{data widths}.

\section{Preliminaries}
In this section, we introduce the preliminaries regarding hyperdimensional computing, including backgrounds of HDC and the process of using HDC in learning tasks.

\subsection{Backgrounds of HDC}
\mypara{Hypervectors} Hypervector (HV) is a type of high-dimensional, holographic vectors with i.i.d. elements~\cite{kanerva2009hyperdimensional}. Assume we have an HV of $n$ dimensions which can be noted as Eq.~\ref{eqn:hv}, where $h_i$ denotes the elements inside the HV. 

In HDC, HVs use their high dimensional space to store different layers of information, thus can represent values, features and even samples. For example, in HDC for image classification tasks, HVs can represent a grayscale value, a pixel or even one image. To establish the dynamic connection between different layers of information in HV, methodologies of aggregating or combing information from HVs such as HDC operations, are therefore necessary.

\begin{equation}
    \vec{H} = \langle h_1, h_2, \dots, h_n\rangle 
    \label{eqn:hv}
\end{equation}

\mypara{Operations}
HVs support three basic operations, addition ($+$), multiplication ($*$) and permutation ($\rho$) as noted in Eq.~\ref{eqn:operation}. Additions and multiplications take two operand HVs as input and perform \textbf{element-wise} operations that add or multiply each element inside the operand HVs index by index. Permutations only take one operand HV and perform \textbf{cyclic shift} over the HV. For all the three operations, the input HVs and the output HVs are in the same dimension.

\begin{equation}
\begin{aligned}
   & \vec{H_p} + \vec{H_q} = \langle h_{p1} + h_{q1}, h_{p2} + h_{q2}, \dots, h_{pn} + h_{qn}\rangle \\
&    \vec{H_p} * \vec{H_q} = \langle h_{p1} * h_{q1}, h_{p2} * h_{q2}, \dots, h_{pn} * h_{qn}\rangle \\
 &   \rho_1(\vec{H}) = \langle h_n, h_1, h_2, \dots, h_{n-1}\rangle 
    \end{aligned}
\label{eqn:operation}
\end{equation}

Addition is used to aggregate parallel features that usually belongs to one modular, while multiplication is used to combine different types of features together to create new features. Permutation is used to reflect spatial or temporal changes in the features. 

\mypara{Similarity Check}
Similarity check is used in HDC for the objective of measuring the similarity $\delta$ of information between different HVs. There are different algorithms to measure similarity such as Euclidean distance and Hamming distance, while in \model, we are using cosine similarity as noted in Eq.~\ref{eqn:cosim}. A higher similarity between two HVs indicates that they share more alike information, or vice versa. 

\begin{equation}
    \delta(\vec{H_p}, \vec{H_q}) = \frac{\vec{H_p} \cdot \vec{H_q}}{||\vec{H_p}||\times||\vec{H_q}||} = \frac{\sum_{i=1}^{n} h_{pi} \cdot h_{qi}}{\sqrt{\sum_{i=1}^{n} {h_{pi}}^2} \cdot \sqrt{\sum_{i=1}^{n} {h_{qi}}^2}}
\label{eqn:cosim}
\end{equation}

\subsection{HDC in Learning Tasks}
\label{sec:hdc_prelim}
HDC in learning tasks features three major phases: \textbf{Training}, \textbf{Inference} and \textbf{Retraining}; all the phases require a fundamental procedure called \textbf{Encoding}.

\mypara{Encoding}
Encoding is the process of mapping input features of one sample to the high-dimensional space available for HDC training, inference and retraining, i.e., building representative HVs of this sample from the fundamental item memory using combinations of HDC operations. Item memory is a type of specially allocated memory during runtime, which stores the bottom layer HVs, or building block HVs that are used to add or multiply for establishing other HVs. To ensure the i.i.d. property, HVs in the item memories are all randomly initialized. 

Process of encoding can be noted as Eq.~\ref{eqn:encode}. Assume we have the $m$-dimensional input features $\vec{F} = \langle f_1, f_2, \dots, f_m \rangle$ for each sample, a set of corresponding item memories $\mathcal{R} = \{\mathcal{R}_1, \mathcal{R}_2, \dots, \mathcal{R}_m\}$ and the combination of HDC operations $E$ determined by the application, the encoded HV $\vec{H}$ is obtained by looking up each feature's corresponding HV in the item memory and then applying them into the HDC operation combination. This encoded HV will subsequently represent the input sample in training, inference and retraining.

\begin{equation}
    \begin{aligned}
    \vec{H} = E(\mathcal{R}, \vec{F}) = E(\mathcal{R}_1[f_1], \mathcal{R}_2[f_2], \dots, \mathcal{R}_m[f_m])
\end{aligned}
\label{eqn:encode}
\end{equation}

\mypara{Training}
Training is the process of aggregating encoded hypervectors sharing the same label together to build the associative memory. Associative memory stores the class HVs, each representing a class in the learning problem. At the beginning of training phase, all the HVs in the associative memory is initialized as zero. Process of training can be denoted as Eq.~\ref{eqn:training}. In the learning problem with $k$ classes, assume we have encoded the HVs $\vec{H^l}$ for each training sample where $l$ means the class label, training process to establish associative memory $\mathcal{A}$ is by adding up hypervectors representing samples from the same class in the training set together.

\begin{equation}
    \begin{aligned}
    \mathcal{A} = & \{ \vec{A^1}, \vec{A^2}, \dots, \vec{A^k}  \} \\
    = & \{ \sum_{}^{}\vec{H^1}, \sum_{}^{}\vec{H^2}, \dots, \sum_{}^{}\vec{H^k} \}
\end{aligned}
\label{eqn:training}
\end{equation}

\begin{figure*}
    \centering
    \includegraphics[width = 1.6\columnwidth]{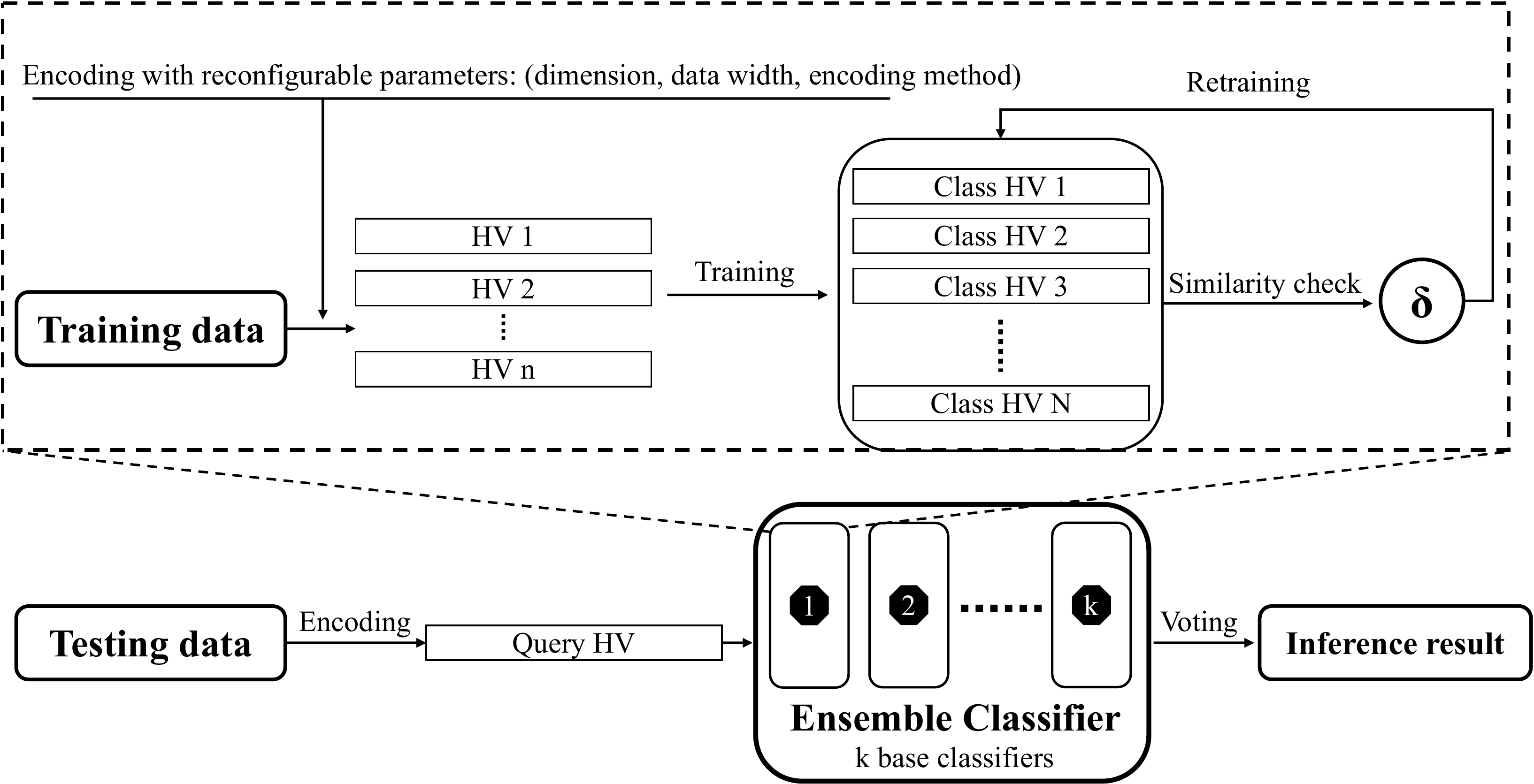}
	\caption{\model~framework}
    \label{fig:framework}
    \vspace{-10pt}
\end{figure*}

\mypara{Inference}
Inference is the process of using the associative memory established in the training phase to determine the class label of an unknown sample. Process of inference can be denoted as Eq.~\ref{eqn:inference}. First, we encode the unknown sample into its representing HV $\vec{H_q}$ referred to as the query HV. Then we perform similarity check between the query HV and each class HV inside the associative memory. As aforementioned, higher similarity means higher common information shared by the two HVs, further indicating that these two HVs are likely from the same class. Therefore, the class of HVs in the associative memory having the highest similarity is determined to be the class of the query HV, namely the predicted label of the sample.

\begin{equation}
    \begin{aligned}
    l = argmax(\{ \delta(\vec{H_q}, \vec{A^1}), \delta(\vec{H_q}, \vec{A^2}), \dots, \delta(\vec{H_q}, \vec{A^k}) \})
\end{aligned}
\label{eqn:inference}
\end{equation}

\mypara{Retraining}
Retraining is the process of fine-tuning the HDC model by correcting erroneous information from the associative memory. The process of retraining can be denoted as Eq.~\ref{eqn:retrain}. Assume we have a query HV $\vec{H_q}$ with class label $l$. If it is predicted wrongly as class $l'$, then we subtract it from the $l'$ class HV $\vec{A^{l'}}$ in the associative memory to remove the erroneous information and add it into the correct $l$ class HV $\vec{A^l}$. By doing this update, the similarity between $\vec{H_q}$ and $\vec{A^{l'}}$ will be reduced while the similarity between $\vec{H_q}$ and $\vec{A^l}$ will be increased to enhance the accuracy of model.
\begin{equation}
    \begin{aligned}
    \vec{A^{l'}} &= \vec{A^{l'}} - \vec{H_q} \\
    \vec{A^{l}} &= \vec{A^l} + \vec{H_q}
\end{aligned}
\label{eqn:retrain}
\end{equation}

\section{\model~Model}
In this section, we describe the process of developing \model~and how to enhance the diversity of the base classifiers in \model~to further improve the performance. Fig.~\ref{fig:framework} illustrates the overview of \model. 
First, we separately train and retrain several different base classifiers using different parameter configurations. These base classifiers are well-trained and we integrate these base classifiers into one ensemble classifier. Secondly, we encode the testing sample into query HV and do the inference on every base classifier. And in phase three, as we collected all the base inference results in phase 2, we employ majority voting to get the ensemble inference result of the ensemble classifier.

\subsection{Base Classifier Development}
In traditional ensemble learning, base classifiers are developed with different initialization settings. In HDC, base classifiers are developed as described in Section~\ref{sec:hdc_prelim}. Each base classifier has different configurations and different randomly generated item memories $\mathcal{R}$. Thus, the training outcome, i.e., the associate memory, will have different class HVs representing the same class. Therefore, for a given query input, the classification output may be different. 

\subsection{Diversity Enhancement of Base Classifier}
To further enhance the diversity of base classifiers, we propose different configurations for base classifiers, including different encoding mechanisms, different dimensions, and different data width for representing the numbers. 

\mypara{Encoding mechanisms:} We use two encoding mechanisms: the Record based encoding and the N-gram based encoding. First of all, the \textbf{Record Encoding} is a general encoding method which mapping every feature vector
$\vec{F} = \langle f_1, f_2, \dots, f_m \rangle$, containing $m$ features, into hyper-dimensional space. The Record encoding method finds the minimum and maximum feature values and converse the range into $p$ feature levels.
In Record encoding method, we assign a set of random orthogonal bipolar level HVs $\vec{H_l}$ = $\langle h_{l1}, h_{l2}, \dots, h_{ln}\rangle$ to every feature levels. Meanwhile, for preserving the position independence of feature values in the feature vector, Record encoding method also assign one set of HVs to each feature values, called base HVs $\vec{H_b}$ = $\langle h_{b1}, h_{b2}, \dots, h_{bn}\rangle$. The Record encoding method use the level HVs to represent each feature value in the feature vector and Base HVs for the position relationship of features values. Then Record encoding employed by linearly combining the level HVs and Base HVs in hyper-dimensional space, the detail of Record encoding shows as Eq.~\ref{eq:record_encoding}. 

\begin{equation}
\vec{H}_{Record} = \sum_{i = 1}^{m} \vec{H}_{li}*\vec{H}_{bi}
\label{eq:record_encoding}
\end{equation}

In Eq.~\ref{eq:record_encoding}, $\vec{H}_{Record}$ is the non-bipolar encoded HV  with D dimensions containing several integer values. Meanwhile, since base HVs are randomly generated in hyper-dimensional space, these HVs are almost mutually orthogonal, which means the cosine similarity between any two base HVs $\vec{H}_{bi}$ and $\vec{H}_{bj}$, we have $\delta$($\vec{H}_{bi}$, $\vec{H}_{bj}$) = 0.

We also use the \textbf{N-gram Encoding} method, in which we employ the locality-based sparse random projection~\cite{BRIC} as our N-gram encoding method. In this method, while we are using the D dimensional HVs, we first extended the length of feature vector from $N$ to D. For instance, in \MNIST~dataset, when we try to encode a feature vector with N = 768 feature values into D = 10000 dimension HVs, we firstly need to attach 13 duplication following the original feature vector. Meanwhile, we generate a random bipolar D dimensional local-hashing HV $\vec{H_s}$ = $\langle h_{s1}, h_{s2}, \dots, h_{sn}\rangle$. To encode the extended feature vector into the hyper-dimensional space, N-gram Encoding method deploy an N-gram sliding window and take the dot product of the extended feature vector and projection vector in this window range. The detailed encoding progress of N-gram Encoding method shows as Eq.~\ref{eq:ngram_encoding}. In this situation, the i-th value of $\vec{H}_{N-gram}$ equals to the dot product of the w feature values from $f_i$ to $f_{i+w-1}$ and w element values from $h_i$ to $h_{i+w-1}$, where $w$ equals to the size of sliding window.

\begin{equation}
\begin{aligned}
& \vec{H}_{N-gram} = \langle v_1, v_2, \dots, v_n\rangle \\
& {v}_i = f_{i}*h_{i} + f_{i+1}*h_{i+1} + \dots + f_{i+w-1}*h_{i+w-1}
\label{eq:ngram_encoding}
\end{aligned}
\end{equation}

\mypara{Dimensions: }We use three different dimension settings for base classifiers, 1000, 5000, and 10000. This brings another diversity aspect for the base classifiers. 

\mypara{Data width: }To represent the internal data of the HDC base classifiers, we use $INT\_8$ and $INT\_16$ data width to further enhance the diversity of base classifiers. 

\subsection{Voting Mechanism}
The inference phase has two steps in the \model. First, we map each testing data into a query HV $\vec{H_q}$ using the same encoding method during training. Then we calculate the cosine similarity of each class HVs with the query HV $\vec{H_q}$ in every base classifier, the inference result is pointed to the class with the highest cosine similarity. Then we finish the inference process in each base classifier following the same routine in step 1. And for step 2, we collect all the base inference results in the ensemble classifier to vote the ensemble inference result. We explored different voting mechanisms to get a better inference result and we tested two different voting strategies, soft voting and hard voting. Hard voting is the majority voting. And for soft voting, since each base classifier gave the inference result by cosine similarity checking, we can sum up all the related cosine distances and rank them in order, where the champion will be selected as the final result. Our test result shows that the hard voting strategy achieved better accuracy in \model. Therefore, we integrate all the base inference results in ensemble classifier and use majority voting to get the ensemble inference result of the corresponding query HV.

\begin{figure*}
    \centering
    
    \subfigure[\MNIST]{
        \includegraphics[width = 0.5\columnwidth]{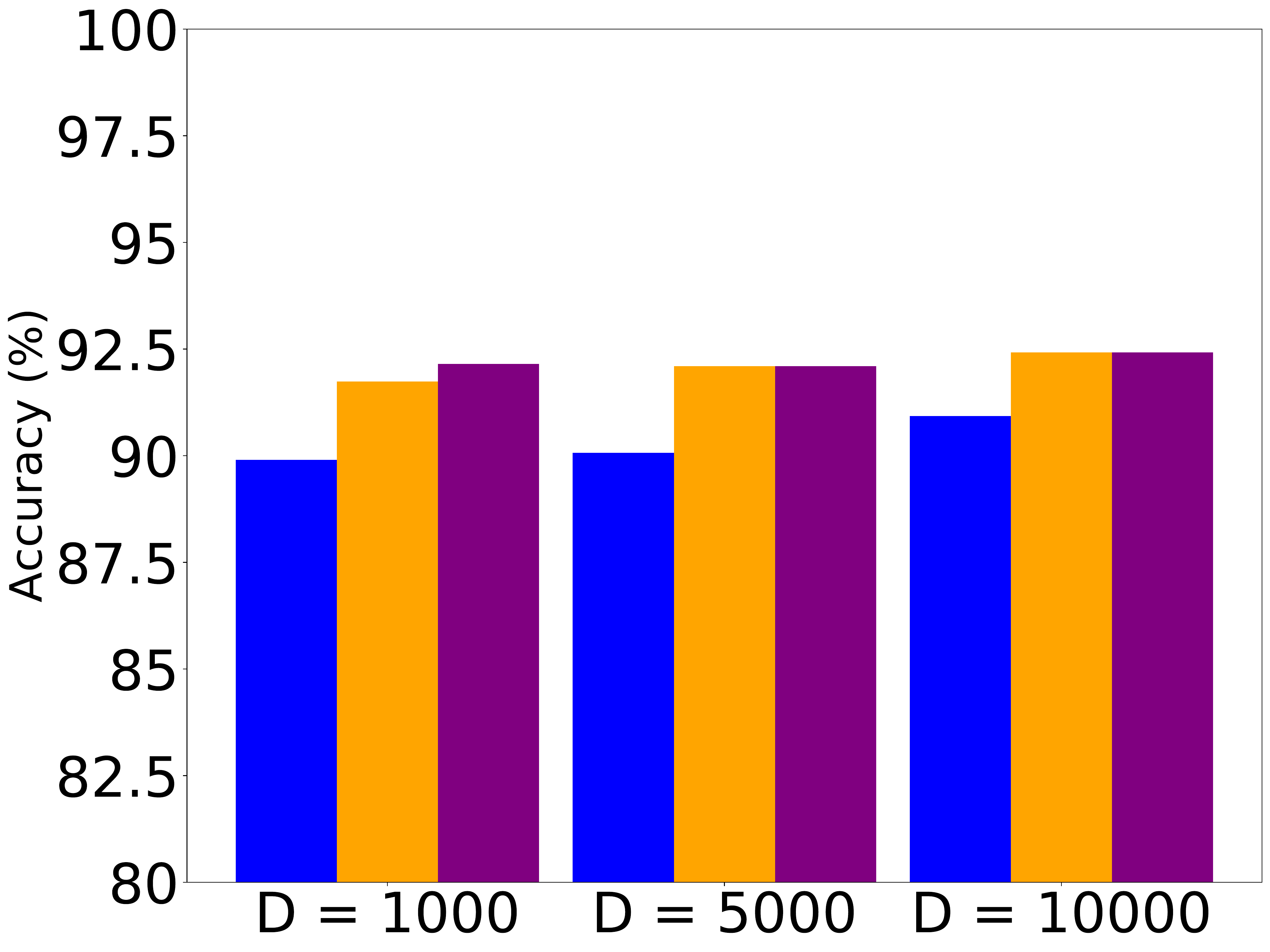}
    }
    \hskip -1em 
    \subfigure[\CARDIO]{
        \includegraphics[width = 0.5\columnwidth]{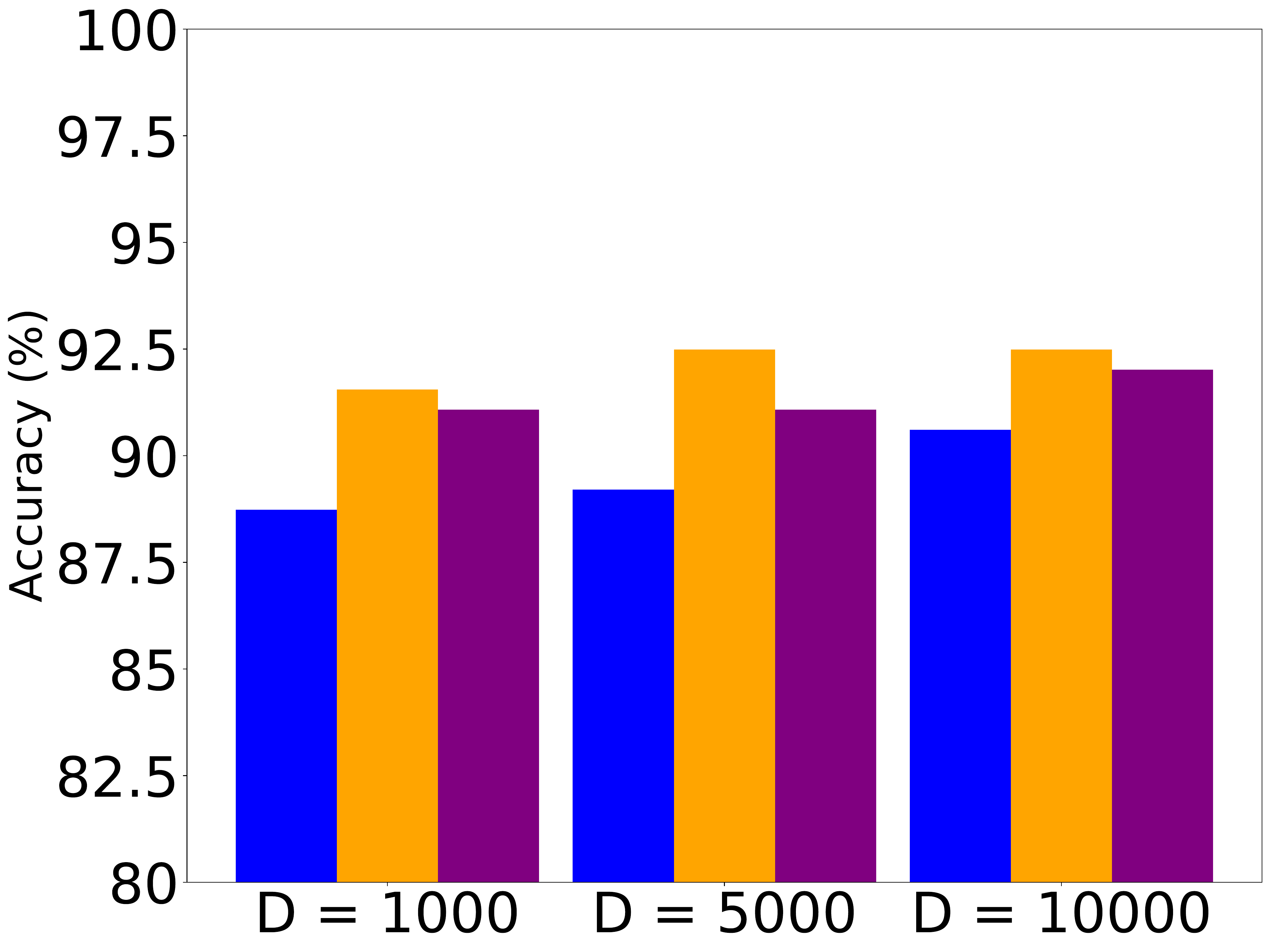}
    }
    \hskip -1em 
    \subfigure[\ISOLET]{
        \includegraphics[width = 0.5\columnwidth]{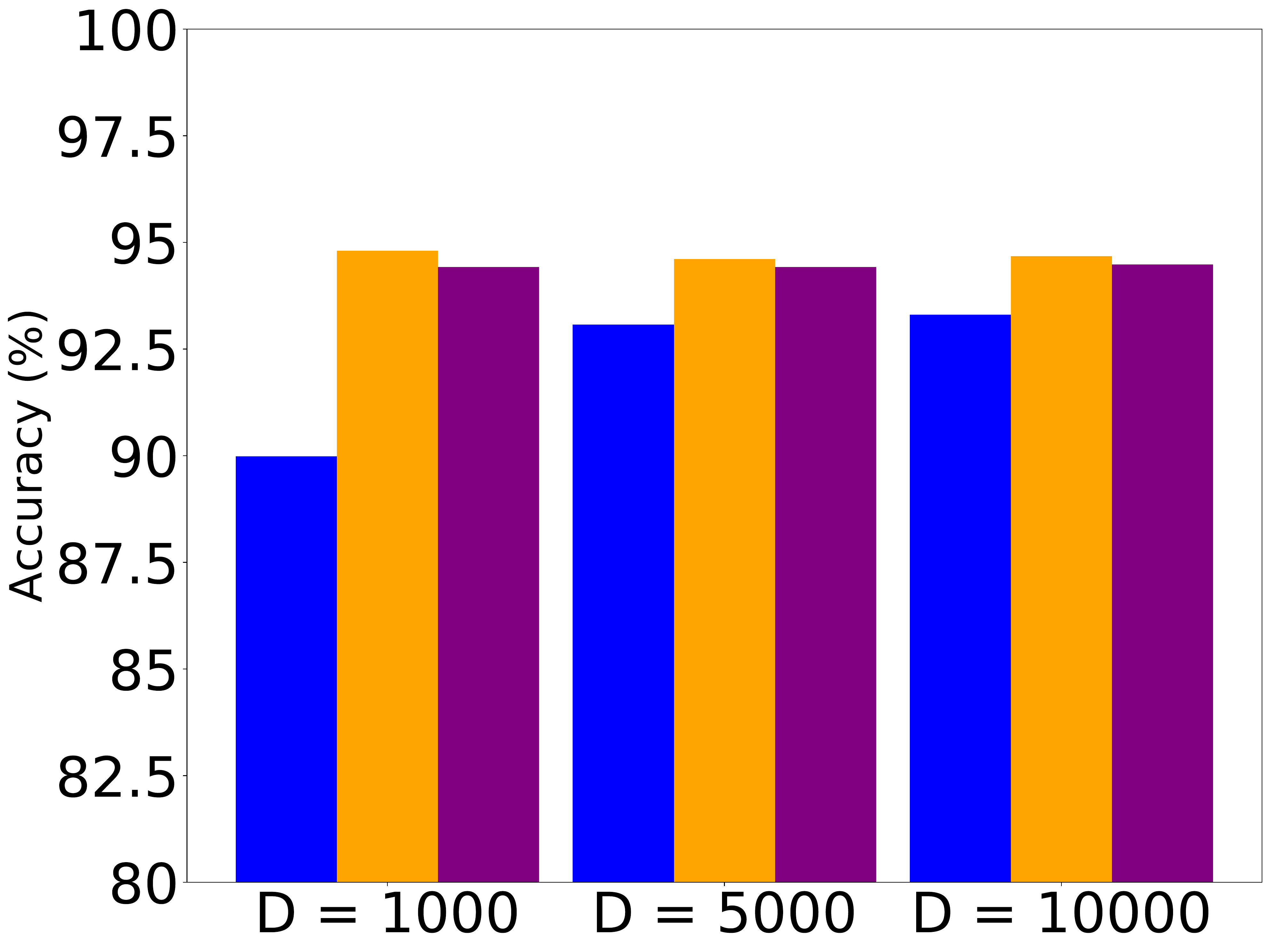}
    }
    \hskip -1em
    \subfigure[\UCIHAR]{
        \includegraphics[width = 0.5\columnwidth]{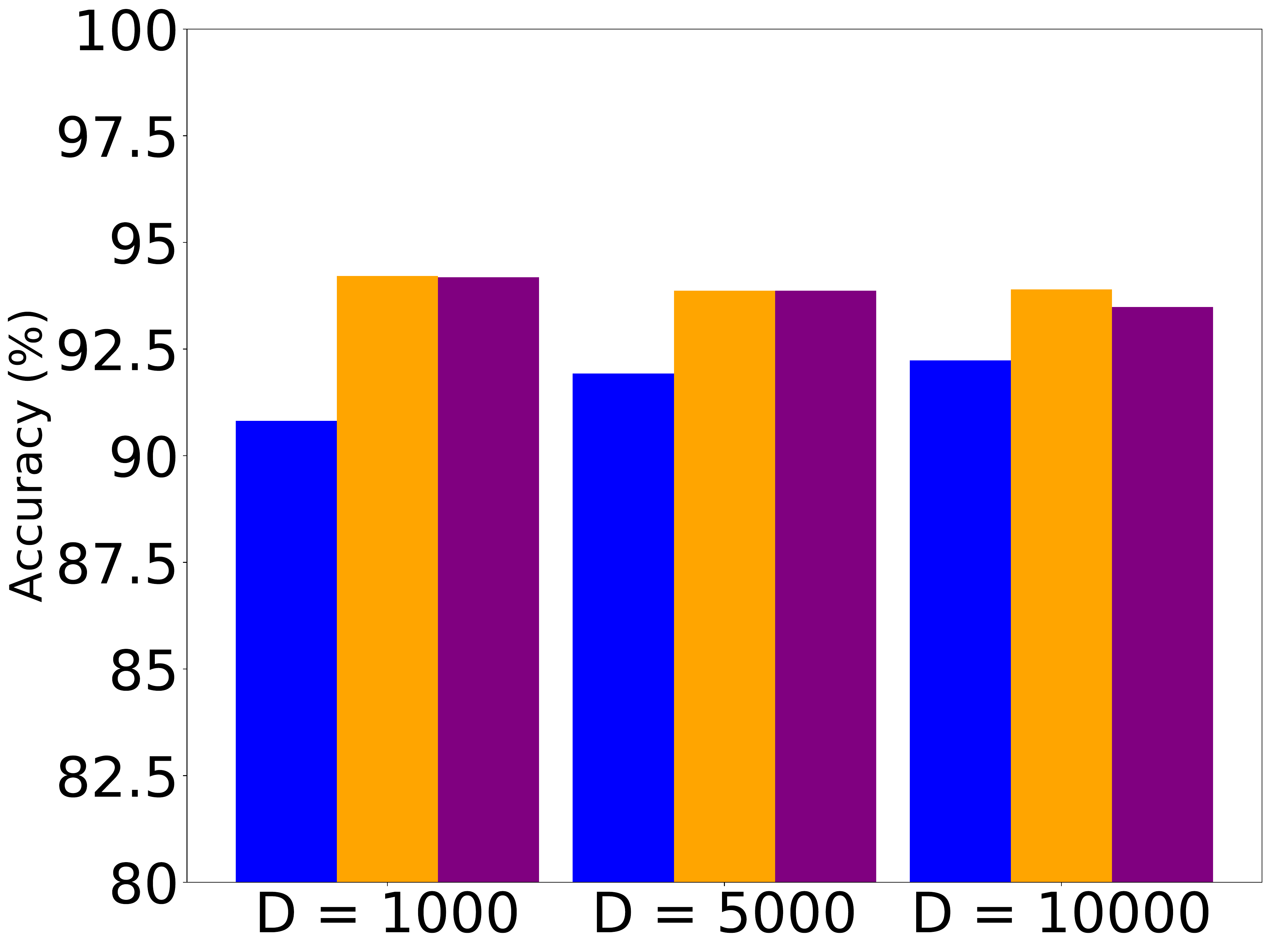}
    }
    \hskip -1em
	
	\includegraphics[width=0.9\columnwidth]{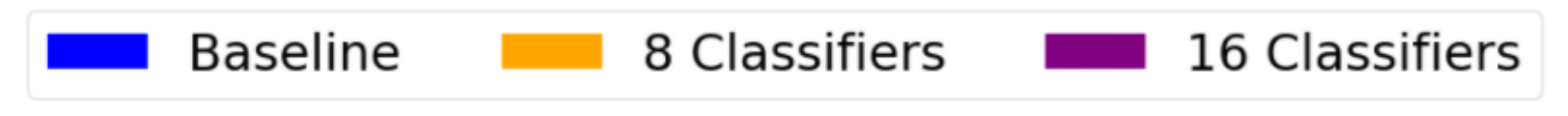}
	\caption{\model~performance under different HV dimension: D = 1000, 5000,  10000}
    \label{fig:dim}
    \vspace{-10pt}

\end{figure*}

\section{Experimental Results}
\begin{figure*}
    \centering
    \subfigure[\MNIST]{
        \includegraphics[width = 0.5\columnwidth]{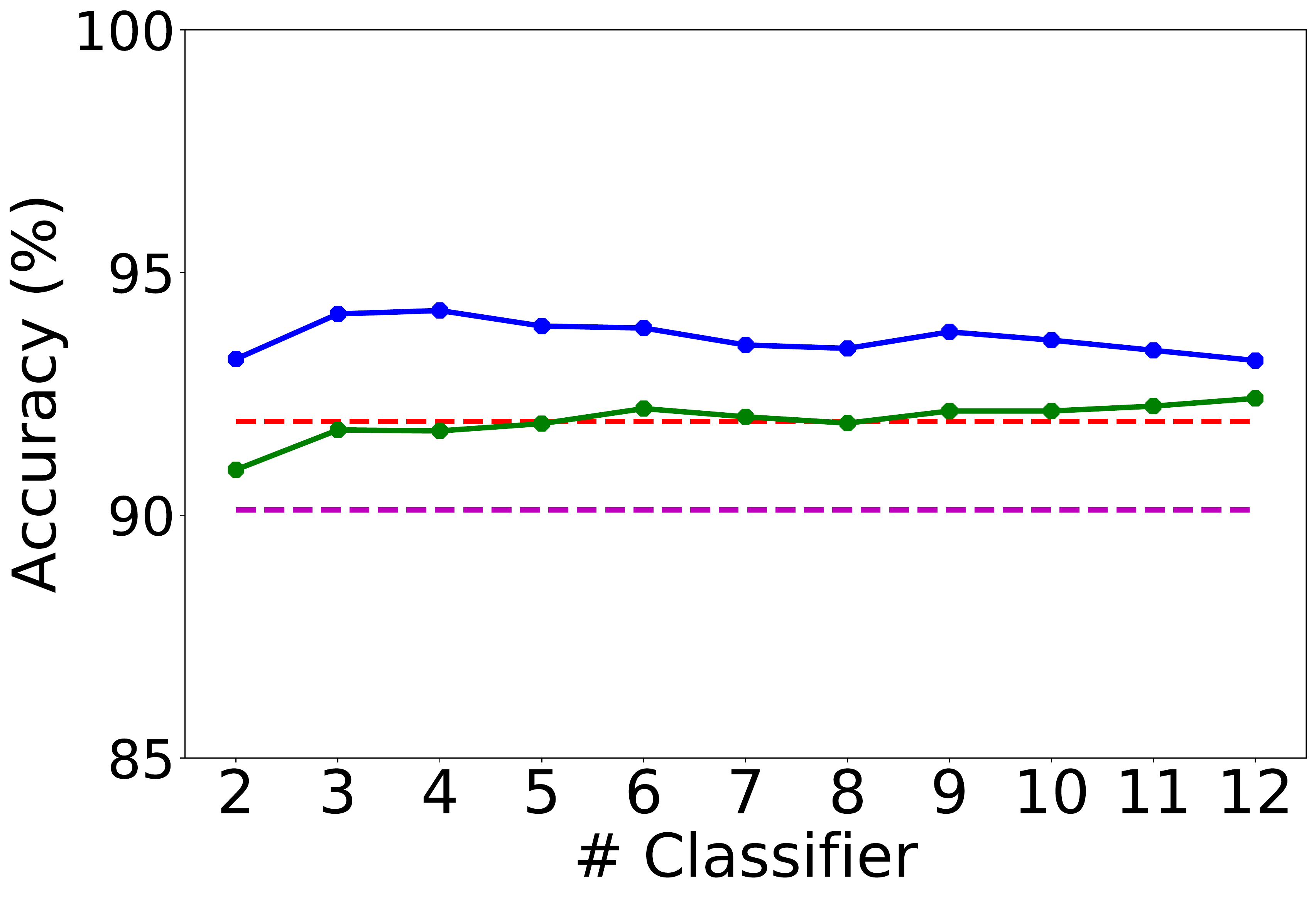}
    }
    \hskip -1em
    \subfigure[\CARDIO]{
        \includegraphics[width = 0.5\columnwidth]{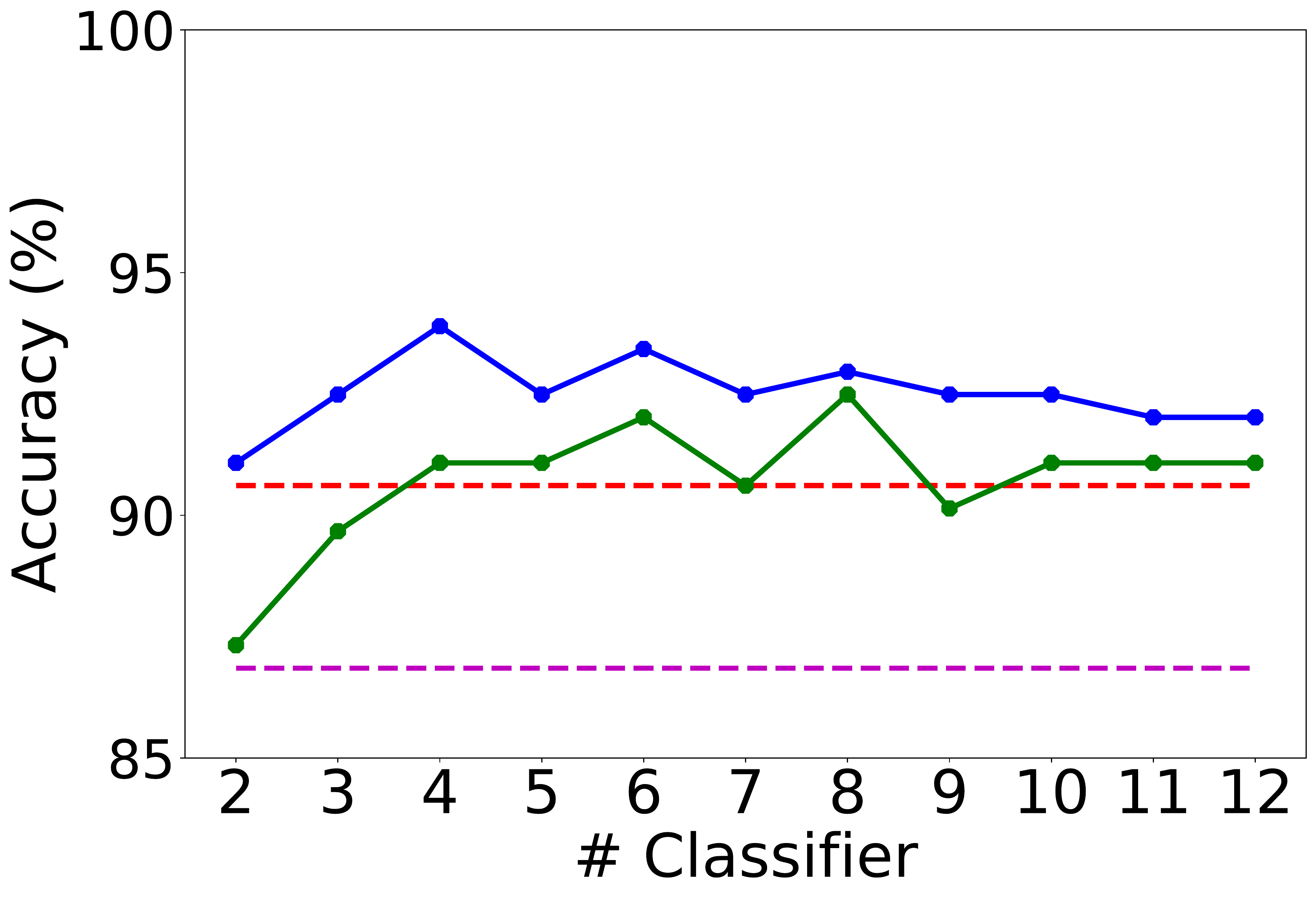}
    }
    \hskip -1em 
    \subfigure[\ISOLET]{
        \includegraphics[width = 0.5\columnwidth]{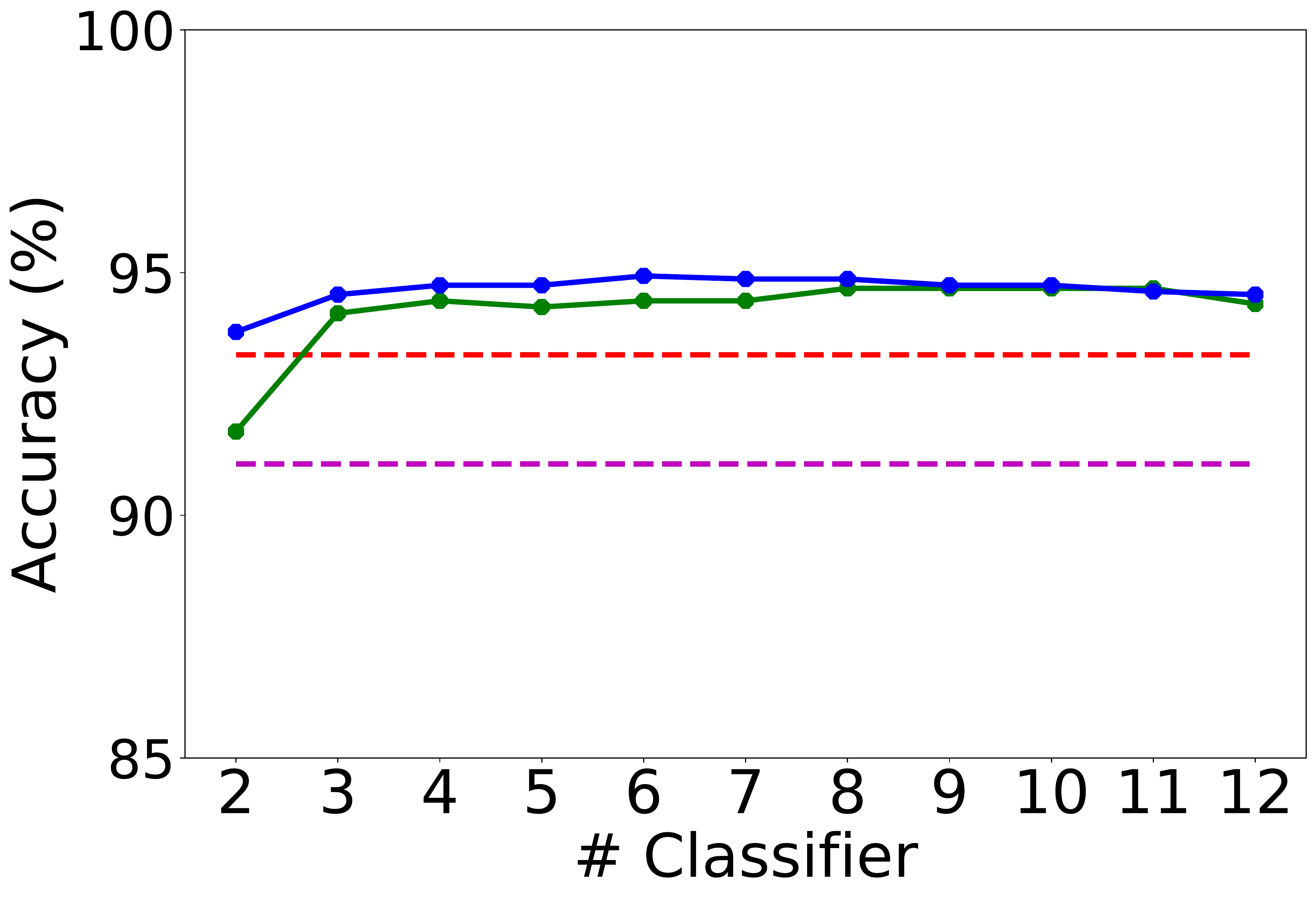}
    }
    \hskip -1em 
    \subfigure[\UCIHAR]{
        \includegraphics[width = 0.5\columnwidth]{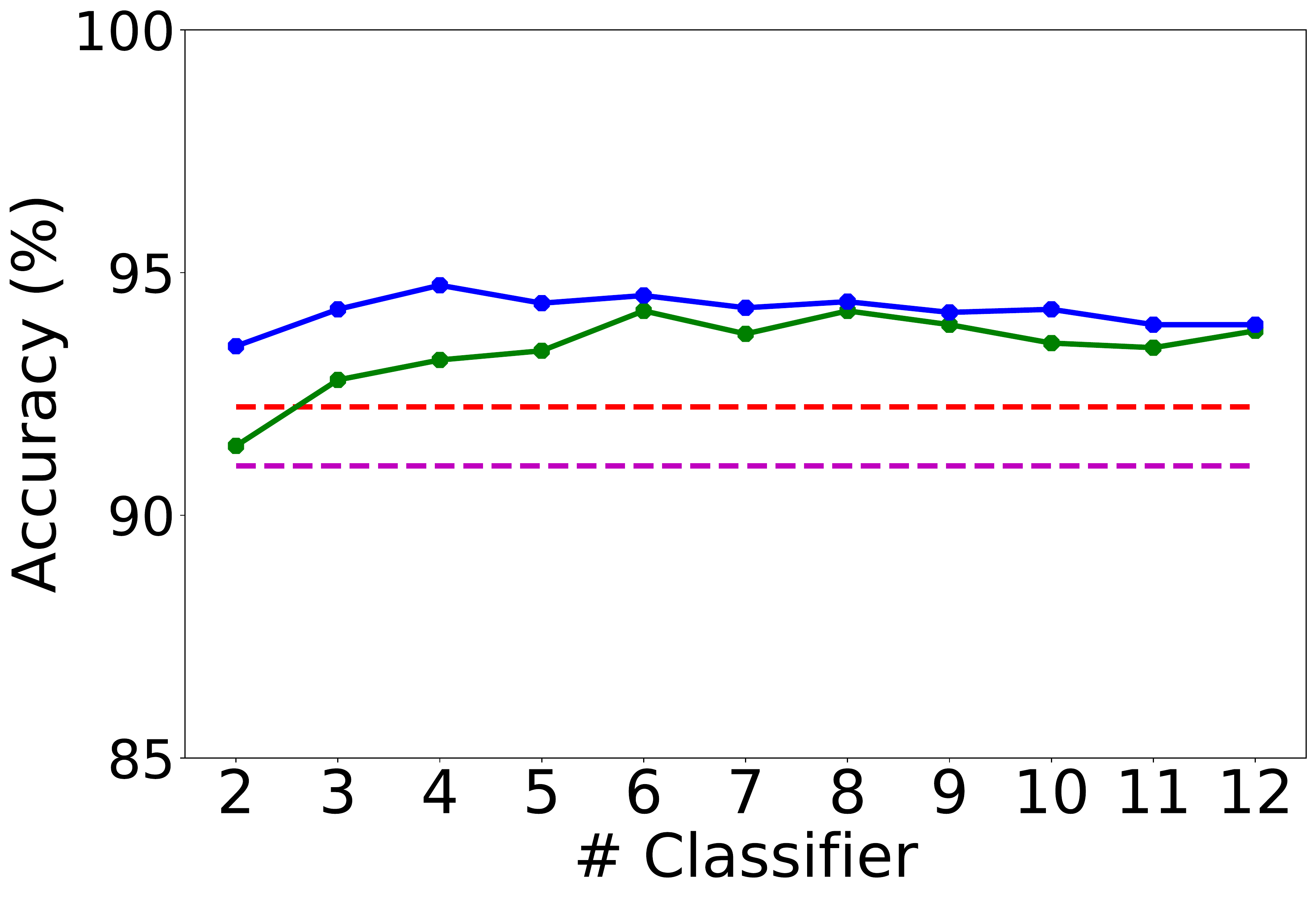}
    }
    \hskip -1em
    \centering
    \includegraphics[width=1.9\columnwidth]{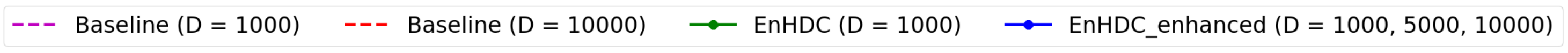}
	\caption{\model~performance under different ensemble size}
    \label{fig:multi}
    \vspace{-10pt}
\end{figure*}

\subsection{Experiment Setup}
We evaluate \model ~using four application domains: speech recognition (\ISOLET~\cite{Dua:2019}), human activity recognition (\UCIHAR~\cite{anguita2013public}), handwritten digits (\MNIST~\cite{lecun1998gradient}), and cardiotocography (\CARDIO~\cite{Dua:2019}). Detail of these four applications are shown below:

\begin{itemize}
\item\MNIST: Handwritten digits Recognition dataset aiming at classifying 10 handwritten digits from 0 to 9. In this application, we use 50000 samples for training and 10000 samples for testing.

\item\CARDIO: Cardiotocography dataset aiming at classifying measurements of fetal heart rate (FHR) and uterine contraction (UC) features into 10 classes (label consensus by obstetricians). It has a total of 2,126 fetal cardiotocograms (CTGs) signals. In this application, we use 1913 samples for training and 213 samples for testing.

\item\UCIHAR: Human activity recognition dataset aiming at recognizing 12 types of human activities. The dataset is built from 30 subjects performing activities of daily living (ADL) with 561 features. In this application, we use 7767 samples for training and 3162 samples for testing.

\item\ISOLET: Speech recognition dataset aiming at recognizing voice audio of the 26 letters of the English alphabet. This dataset contains 150 subjects speaking the name of each alphabet letter twice. In this application, we use 6238 samples for training and 1559 samples for testing.

\end{itemize}

\subsection{Accuracy Improvement}
Fig.~\ref{fig:dim} and Fig.~\ref{fig:multi} present the accuracy comparison between different configurations of \model. The baseline is one HDC classifier while the \model~employing several different base classifiers. According to our experiment result, we can observe several important facts.

First as shown in Fig.~\ref{fig:dim}, \model~contains 8 and 16 base classifiers with different encoding methods (Record and N-gram encoding). To evaluate the performance of \model, we compare 3 models under the same dimensionality setting across $D = 1000,5000,10000$, \model~is showing consistent higher accuracy than baseline models. When \model~uses 8 base classifiers, the average improvement is 3.2\% over all applications.

Second, normally HDC requires a high dimensionality such as $D=10000$ to achieve satisfying performance. However, with \model, we can even achieve higher accuracy under lower dimensionality than that of baseline model with higher dimensionality. Actually, across all the applications, \model ~with 8 classifiers under $D=1000$ presents higher accuracy than baseline model under $D=10000$. The average improvement is 1.37\%.

Third, as shown in Fig.~\ref{fig:multi}, the number of base classifiers have a notable impact on the accuracy. Without the loss of generality, our experiment features different ensemble sizes, starting from two base classifiers to twelve base classifiers. As the green line shown in Fig.~\ref{fig:multi}, the accuracy of \model ~increases by adding more classifiers but comes to the vertex when using eight base classifiers for most applications. After this, the accuracy improvement is saturated. This is consistent with the ensemble theory in~\cite{lessismore}, where the performance of ensemble learning algorithms cannot constantly increase by adding an infinite amount of base classifiers. The accuracy will peak during the progress of increasing the number of classifiers, and after this peak value, the overall accuracy cannot have an obvious improvement. 

Last but not least, the diversity enhancement can further improve the accuracy of \model~as shown in Fig.~\ref{fig:multi}. \model\_enhanced classifier is the \model ~classifier with enhanced diversity by varying encoding mechanisms (Record encoding, N-gram encoding), dimensions ($D=1000, 5000, 10000$), and data width ($INT\_8$, $INT\_16$). This figure shows that the \model\_enhanced classifier can further improve the accuracy of \model~classifier by 1.2\% on average across all applications, which is 4.4\% improvement over baseline HDC model.  

\subsection{Memory Reduction}
With ensemble learning, we are able to reduce the dimensionality requirement for HDC classifiers. Typically, HDC is required to have a high dimensionality, e.g., 10000, to achieve a satisfying performance. For example, for \CARDIO~dataset in Fig.~\ref{fig:dim}, the baseline HDC classifier with 10000 dimensions has 1.9\% higher accuracy than the baseline HDC classifier with 1000 dimensions.  

However, with ensemble learning, we can see that \model ~with just 1000 dimensions is able to achieve similar level or even surpass the accuracy of 10000-dimension baseline HDC classifiers across all applications. This can result in a smaller model size. For example, with 8 base classifiers with $D=1000$ dimensions, this can reduce 20\% model size compared to a baseline classifier with $D=10000$ dimensions. For \MNIST~dataset with 10 classes, we have a baseline HDC model with $D=10000$ and $INT\_8$ data width whose model size is $8~bits * 10000~dimensions * 10~classes$ = 800Kb and \model ~with base classifiers with $D=1000$ and $INT\_8$ whose model size is $8~bits * 1000~dimensions * 10~classes * 8~classifiers$ = 640Kb. The overall model size reduction is 160Kb for \MNIST. Meanwhile, in \UCIHAR~dataset with 12 classes, we have one 960Kb baseline model with $D=10000$ and \model~with eight 96Kb base classifiers with $D=1000$, pointing out the model size reduction for \UCIHAR ~is 180Kb.

\section{Conclusion}
This paper presents \model, the first ensemble HDC classifier. As an ensemble learning model, \model ~employs different base classifiers under different HV dimensions, different data widths, and different encoding methods. In \model, base classifiers are individually trained and retrained and \model ~applies the majority voting to generate the final inference result. We designed several experiments to evaluate the performance of \model ~under different HV dimensions and different number of classifiers. By evaluating \model ~on four practical applications, we show that \model ~can achieve higher accuracy and can reduce model size compared to baseline HDC classifiers while providing the same or even better classification accuracy. Meanwhile, by increasing the diversity of base classifiers in \model, the classification accuracy has an enhanced improvement compared to the original \model ~model. This paper presents the first effort in using an ensemble learning method in HDC for boosting the performance and opens the door for this potential research direction. Our future work will consider using a cascading method and more sophisticated ensemble learning algorithms such as boosting.

\bibliography{EnHDC}

\end{sloppypar}
\end{document}